\RequirePackage{amsthm}
\documentclass[sn-apa]{sn-jnl}

\usepackage{graphicx}%
\usepackage{multirow}%
\usepackage{amsmath,amssymb,amsfonts}%
\usepackage{amsthm}%
\usepackage[title]{appendix}%
\usepackage{xcolor}%
\usepackage{textcomp}%
\usepackage{manyfoot}%
\usepackage{booktabs}%
\usepackage{algorithm}%
\usepackage{algorithmicx}%
\usepackage{algpseudocode}%
\usepackage{listings}%
\usepackage{array} %
\usepackage{lmodern}%
\usepackage{ragged2e}

\theoremstyle{thmstyleone}%
\theoremstyle{thmstyletwo}%

\theoremstyle{thmstylethree}%

\begin{document}

\title[Article Title]{AdvBlur: Adversarial Blur for Robust Diabetic Retinopathy Classification and Cross-Domain Generalization}


\author*[1]{\fnm{Heethanjan Kanagalingam} \sur{Author}}\email{heethanjanheetha@gmail.com}
\equalcont{These authors contributed equally to this work.}

\author[1]{\fnm{Thenukan Pathmanathan} \sur{Author}}\email{thenukanpt@gmail.com}
\equalcont{These authors contributed equally to this work.}

\author[2]{\fnm{Mokeeshan Vathanakumar} \sur{Author}}\email{accmokee30@gmail.com}
\equalcont{These authors contributed equally to this work.}

\author[3]{\fnm{Tharmakulasingam Mukunthan}
\sur{Author}}\email{mukunthan@eng.jfn.ac.lk}

\affil*[1]{\orgdiv{Department of Electronic and Telecommunication Engineering}, \orgname{University of Moratuwa}, \orgaddress{
  \street{Katubedda}, 
  \city{Moratuwa}, 
  \postcode{10400}, 
  \state{Western Province}, 
  \country{Sri Lanka}
}}
\affil*[2]{\orgdiv{Department of Biomedical Engineering}, \orgname{University of Melbourne}, \orgaddress{
  \street{Parkville}, 
  \city{Melbourne}, 
  \postcode{3010}, 
  \state{Victoria}, 
  \country{Australia}
}}

\affil[3]{\orgdiv{Department of Electrical and Electronic Engineering}, \orgname{ University of Jaffna}, \orgaddress{
  \street{Ariviyal Nagar}, 
  \city{Killinochchi}, 
  \postcode{44000}, 
  \state{Northern Province}, 
  \country{Sri Lanka}
}}


\abstract
{Diabetic retinopathy (DR) is a leading cause of vision loss worldwide, yet early and accurate detection can significantly improve treatment outcomes. While numerous Deep learning (DL) models have been developed to predict DR from fundus images, many face challenges in maintaining robustness due to distributional variations caused by differences in acquisition devices, demographic disparities, and imaging conditions. This paper addresses this critical limitation by proposing a novel DR classification approach, a method called \textbf{AdvBlur}. Our method integrates adversarial blurred images into the dataset and employs a dual-loss function framework to address domain generalization. This approach effectively mitigates the impact of unseen distributional variations, as evidenced by comprehensive evaluations across multiple datasets. Additionally, we conduct extensive experiments to explore the effects of factors such as camera type, low-quality images, and dataset size. Furthermore, we perform ablation studies on blurred images and the loss function to ensure the validity of our choices. The experimental results demonstrate the effectiveness of our proposed method, achieving competitive performance compared to state-of-the-art domain generalization DR models on unseen external datasets.
\keywords{Diabetic Retinopathy, Fundus Images, Blur, Domain Generalization}}



\maketitle

\section{Background}

Diabetic retinopathy (DR) has become one of the leading causes of blindness worldwide, particularly among working-age adults. According to the International Diabetes Federation (IDF), over 500 million individuals globally are affected by diabetes, and nearly one-third of them are expected to develop some form of DR during their lifetime~\cite{IDF2021,world2019world}. This condition arises as a complication of diabetes, where prolonged high blood sugar levels damage the retinal blood vessels, leading to visual impairment and, eventually, blindness if untreated.

The history of DR dates back to the 1850s when Eduard Jaeger and Albert von Graefe first described visible retinal changes in diabetic patients~\cite{jaeger1856beitr, von1858ueber}. In 1872, Edward Nettleship provided definitive evidence of DR using histopathological images~\cite{nettleship1873oedema}. The development of fluorescein angiography in the mid-20th century facilitated a more detailed understanding of DR, leading to establishing the Airlie House classification system for DR~\cite{Benson2021Nov}.

\subsection{Pathophysiology and classification of DR}

DR primarily affects the retina, the light-sensitive layer of tissue in the back of the eye responsible for converting light into neural signals sent to the brain. The condition begins with damage to the small blood vessels (capillaries) in the retina due to chronic hyperglycemia. This damage indicates microaneurysms, the earliest visible lesions in DR, caused by the weakening of the capillary walls~\cite{Patel2021, li2020domain}. In more advanced stages, capillary closure leads to retinal ischemia, causing the release of vascular endothelial growth factor (VEGF), which promotes the growth of new, fragile blood vessels. This neovascularization, characteristic of proliferative diabetic retinopathy (PDR), often results in vitreous hemorrhages and tractional retinal detachment, significantly impairing vision~\cite{gulshan2016development}.

DR is categorized into two primary stages: non-proliferative diabetic retinopathy (NPDR) and PDR \cite{kalyani2023diabetic}. NPDR represents the early stage of the disease, characterized by microaneurysms, intraretinal hemorrhages, and lipid exudates~\cite{Patel2021}. As the condition progresses, capillary occlusion and ischemia become evident, leading to more severe signs. PDR, the advanced stage, is marked by neovascularization and the potential for complications such as vitreous hemorrhage and retinal detachment, which can cause permanent blindness~\cite{tsin2018early}.

The Early Treatment Diabetic Retinopathy Study (ETDRS) grading system is commonly used to classify DR severity into five levels: no DR, mild NPDR, moderate NPDR, severe NPDR, and PDR~\cite{early1991early}. This classification guides treatment strategies, which may include laser photocoagulation, intravitreal injections, or surgical interventions~\cite{chakrabarti2012diabetic}. Accurate classification of DR stages is essential for effective management, and it is here that artificial intelligence (AI)-based diagnostic tools are making significant strides by improving sensitivity and specificity~\cite{gulshan2016development,li2020domain}.

\subsection{AI applications in DR diagnosis}

The journey of computer-aided DR diagnosis began in the 1980s and 1990s with the advent of computer-aided diagnosis (CAD) systems. These systems primarily relied on handcrafted features to detect abnormalities such as microaneurysms, hemorrhages, and exudates in fundus images \cite{ spencer1996}. Early methods used mathematical morphology and simple rule-based algorithms for feature extraction, often combined with statistical classifiers like k-nearest neighbors (k-NN) and support vector machines (SVMs) \cite{niemeijer2007}. While these methods provided a proof of concept, their performance was limited by their reliance on manual feature engineering and sensitivity to variations in imaging conditions.

In the late 2000s and early 2010s, AI emerged as a major part of medical diagnostics, particularly in addressing the early detection and management of diseases such as DR. Leveraging advanced machine learning (ML) methodologies, including deep learning (DL), AI systems have demonstrated the ability to analyze complex medical images with high precision and efficiency~\cite{abramoff2018pivotal, miotto2018deep, ting2019artificial}. Techniques such as random forests and ensemble learning were used to combine multiple hand-made features to improve classification performance \cite{antal2012, quellec2008}.

One significant advantage of AI-driven diagnostic tools is their ability to augment clinical workflows by reducing the workload of ophthalmologists, particularly in high-volume or resource-constrained settings. Such tools are especially valuable in remote regions, where access to specialized healthcare services remains limited~\cite{kermany2018identifying, grzybowski2020artificial}.

The introduction of DL, particularly CNNs, in the early 2010s marked a major shift in DR diagnosis. CNNs allowed models to automatically learn hierarchical features from raw pixel data, eliminating the need for manual feature engineering \cite{lecun2015,gulshan2016development} demonstrated the first large-scale application of CNNs in DR diagnosis, achieving sensitivity and specificity comparable to ophthalmologists using a dataset of over 100,000 images. This study catalyzed a wave of research into DL applications for medical imaging \cite{kermany2018, ting2017}.

With their ability to capture global contextual information, transformers have further enhanced the performance of AI systems in retinal image analysis. Additionally, by integrating temporal data from patient records, AI systems can predict disease progression, facilitating personalized treatment strategies and timely intervention ~\cite{li2020domain, guan2021domain}.

To address the challenge of limited labeled datasets in medical imaging, transfer learning became a popular approach. Pre-trained models, such as VGG and ResNet, were fine-tuned on retinal images, enabling efficient learning with smaller datasets \cite{mutawa2023transfer, aiche2022transfer}. Multi-task learning frameworks further enhanced the utility of AI systems by enabling simultaneous DR grading, macular edema detection, and lesion segmentation \cite{foo2020multi, tang2021multitask}.

Despite these advancements, several challenges must be addressed to ensure the robust and equitable deployment of AI in clinical practice. Ethical considerations, including data privacy, bias mitigation, and equitable access to AI tools, remain critical concerns~\cite{beede2020human}. Additionally, model interpretability plays a vital role in fostering clinician trust and ensuring transparent decision-making~\cite{borys2023explainable}.

Domain generalization is another significant challenge, as AI models trained on specific datasets may underperform when applied to diverse populations or imaging devices~\cite{das2022critical}. Enhancing domain generalization could improve model performance on unseen datasets, which is essential for ensuring reliable and fair clinical applications.

\subsection{Domain generalization for DR}

Domain generalization is a critical research area that aims to develop models capable of generalizing to unseen domains during inference without requiring access to target domain data during training. The concept of domain generalization was first introduced by \citet{blanchard2011generalizing}, addressing the limitations of domain adaptation by enabling models to generalize to novel domains directly.  Early advancements primarily focused on learning invariant representations across domains, such as \citet{muandet2013domain}, which proposed a domain-invariant component analysis framework to minimize domain discrepancy, and \citet{ghifary2015domain}, introduced multi-task autoencoders to enhance feature generalization. These foundational works paved the way for exploring domain generalization in more complex settings \cite{volpi2018generalizing, dou2019domain}. 

Over time, researchers extended domain generalization techniques to include data augmentation, meta-learning, and regularization-based approaches. \citet{volpi2018generalizing} introduced a data augmentation strategy leveraging synthetic data generation, while \citet{li2018learning} proposed a meta-learning approach to enhance adaptability to unseen domains. \citet{dou2019domain} incorporated specialized loss functions to encourage domain-invariant feature learning. These advancements collectively expanded the applicability of domain generalization to real-world problems, particularly in high-stakes domains, as highlighted by \citet{wang2022generalizing}.

Domain generalization has been widely adopted in medical imaging to address variability in imaging protocols, devices, and patient populations across institutions. Studies have shown its effectiveness in handling domain shifts across imaging modalities, including brain tumor segmentation and chest X-ray classification \cite{kundu2021generalize, guan2021domain, khoee2024domain, wang2022generalizing}. In DR classification, domain generalization helps mitigate variations in fundus imaging due to differences in camera settings, illumination, and patient demographics, ensuring more robust and consistent performance across clinical settings \cite{gulshan2016development, ting2017development, lyu2022aadg}.

In 2022, Atwany and Yaqub introduced the DRGen framework to tackle domain generalization challenges in DR classification \cite{atwany2022drgen}. Their approach incorporated a weight-averaging strategy at specific training iterations and a gradient covariance reduction loss. Evaluated using a leave-one-dataset-out strategy across four fundus imaging datasets, DRGen demonstrated significant improvements in generalization performance. Building upon these advancements,\cite{chokuwa2023generalizing} explored the use of variational autoencoders (VAEs) to disentangle latent representations in fundus images. By separating domain-invariant content from domain-specific noise, their approach outperformed contemporary methods on diverse DR datasets.

In 2023, a method was proposed leveraging Contrastive Language–Image Pre-training (CLIP) for domain generalization in DR classification \cite{baliah2023exploring}. They introduced a multi-modal fine-tuning strategy, Context Optimization with Learnable Visual Tokens (CoOpLVT), which conditioned models on visual features, resulting in a 1.8\% F1-score improvement compared to baseline approaches.
Causality-inspired frameworks have also shown promise in addressing domain shift challenges. \cite{wei2024caudr} introduced CauDR, which incorporated do-operations from causal inference into its architecture to remove spurious correlations caused by dataset biases. This method was accompanied by the 4DR benchmark, which evaluates domain generalization scenarios in medical imaging. 

In 2024, a self-distillation technique for vision transformers (ViT) was proposed to enhance domain generalization performance \cite{galappaththige2024generalizing}. By softening one-hot predictions via adaptive convex combinations, this method improved ViT's generalization capabilities on unseen distributions in DR classification. 

In 2024, \cite{radr2024} presented a framework called RADR, which employed domain-adversarial training to achieve robust DR severity classification. RADR incorporated camera-specific metadata, utilizing the camera labels provided by \cite{yang2020residual}, to align features across domains. While this improved robustness, reliance on such metadata poses challenges for scalability in diverse clinical environments. Additionally, RADR leveraged quality control labels from \cite{fu2019evaluation} to mitigate issues related to low-quality images.

 Prior to our work, RADR set the benchmark in DR severity classification with domain generalization. However, our model surpasses RADR in domain generalization while relying solely on the fundus image dataset, without incorporating external metadata. Furthermore, our experiments demonstrate that explicit quality control labeling is not essential for achieving strong performance, reinforcing the adaptability of our approach across varied clinical settings.

\subsection{Contributions}

This paper introduces a novel method, called AdvBlur, to address the challenges of DR diagnosis and domain generalization. By eliminating the reliance on camera-specific information, our approach leverages fundus images irrespective of their source. This methodology aims to enhance the robustness of DR diagnosis across diverse domains.  Our key contributions are as follows:

\begin{enumerate}
    \item We propose a robust Adversarial blurred image integration technique for cross-domain performance in DR diagnosis. -AdvBlur
    \item A novel combined loss function idea has been introduced.
    \item Extensive experiments on diverse datasets demonstrate the effectiveness of our approach, and the ablation studies validated our novelty further. 
\end{enumerate}

The rest of this paper is organized as follows: Section \ref{sec2} details the proposed methodology; Section \ref{sec3} shows experiments and the results; Section \ref{sec4} presents ablation works; and Section \ref{sec5} concludes the work with future directions.

\section{Proposed Method}\label{sec2}
\begin{figure}[ht!]
    \centering
    \includegraphics[width=0.9\textwidth]{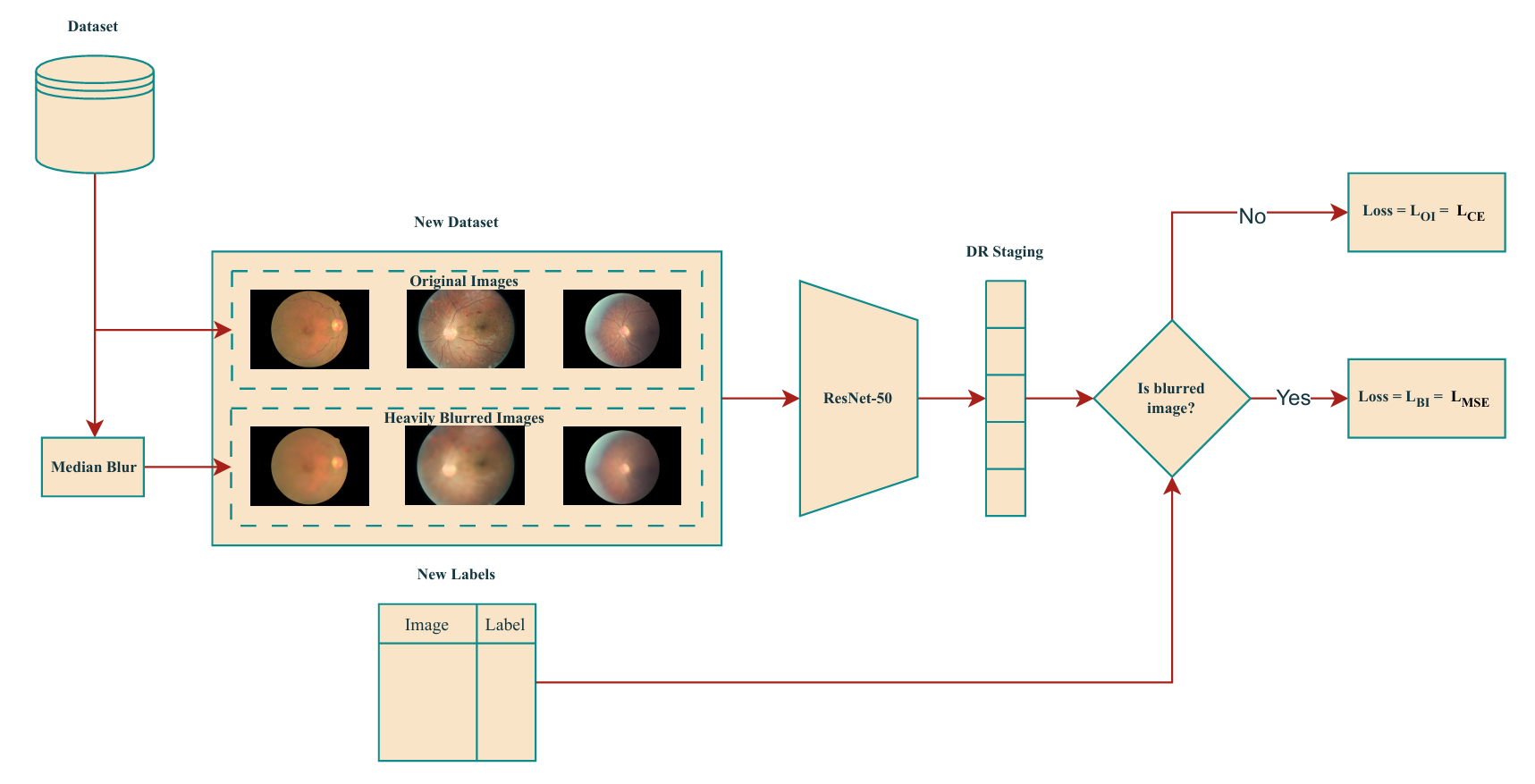}
    \caption{AdvBlur-Proposed methodology. As shown in the diagram, a new dataset is prepared by adding heavily blurred versions of the original images. These blurred images are labeled as Class 6. During training, classification is performed, and the loss function is applied based on the image label. If the image is blurred, the loss function \(L_{\text{BI}}\) (blurred image loss) is used. If the image is original, the loss function \(L_{\text{OI}}\) (original image loss) is applied.}
    \label{fig:architecture}
\end{figure}

In this section, we describe the proposed training strategy for DR classification, the custom loss function introduced to enhance domain generalization, and the integration of heavily blurred images to improve model robustness by guiding the model on what features should not be used for classification. Figure \ref{fig:architecture} shows the overall pipeline of AdvBlur in detail. 

\subsection{Dataset and preprocessing}

The dataset used for training includes fundus images with the corresponding severity labels of DR. The model is trained using the EyePACS dataset, a publicly available collection of 88,702 color fundus eye images  \cite{diabetic-retinopathy-detection}. These images are classified into five classes, corresponding to the level of DR severity. For the evaluation purposes, we used the Messidor-1 \cite{decenciere2014feedback} , Messidor-2 \cite{abramoff2013automated}, and APTOS  \cite{aptos2019-blindness-detection} datasets. The details of all the datasets have been added in Table: \ref{tab:dataset}.

\begin{table}[ht!]
\centering
\caption{Details of EyePACS, Messidor-1, Messidor-2, and APTOS 2019 datasets.}
\begin{tabular}{|>{\centering\arraybackslash}m{0.25\linewidth}|>{\centering\arraybackslash}m{0.15\linewidth}|>{\centering\arraybackslash}m{0.15\linewidth}|p{0.45\linewidth}|}
\hline
\textbf{Dataset} & \textbf{Total Images} & \textbf{Number of Labels} & \textbf{Description} \\ 
\hline
EyePACS \cite{diabetic-retinopathy-detection}& 88,702& 5 & A large dataset of retinal fundus images used for DR detection, notably in Kaggle’s DR competition. \\ 
\hline

Messidor-1 \cite{decenciere2014feedback}& 1,200 & 4 & Contains retinal images for evaluating DR levels, widely utilized in research on automated detection and classification. \\ 
\hline

Messidor-2 \cite{abramoff2013automated} & 1748& 5& An extension of Messidor-1 with similar labeling for DR severity levels, developed for benchmarking algorithms. \\ 
\hline

APTOS 2019 \cite{aptos2019-blindness-detection} & 3662& 5 & Fundus images used in the APTOS 2019 Blindness Detection competition to detect DR severity. \\ 
\hline

\end{tabular}

\label{tab:dataset}
\end{table}

To address domain generalization issues and prevent the model from relying on spurious correlations, we introduce a sixth class comprising heavily blurred versions of the original images. These blurred images are generated using a strong median blur to obscure important retinal features, effectively teaching the model to disregard non-informative visual patterns during classification. Here, the kernal size of 151 had been used for the median blur image generation. The blurred image integration will act as a form of adversarial noise, forcing the model to discard domain-specific high-frequency features in the DR classification. 

\begin{figure}[ht!]
    \centering
    \includegraphics[width=0.8\textwidth]{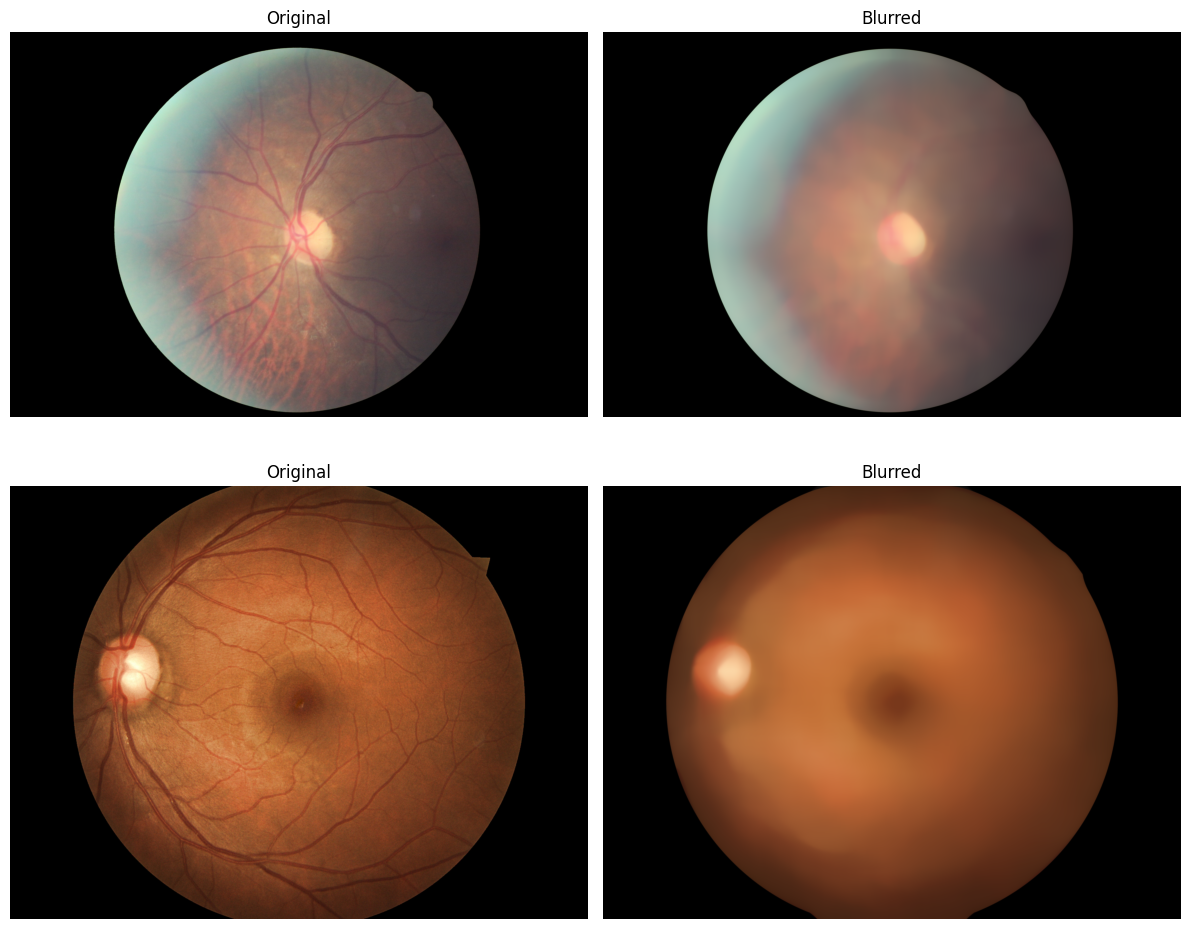}
    \caption{Examples of original (left) and blurred (right) fundus images. The blurring is applied using a median blur with a kernel size of 151.}
    \label{fig:blurred_example}
\end{figure}

Figure \ref{fig:blurred_example} shows examples of original and blurred fundus images used in this study. The blurred images are intended to guide the model on what features should be ignored during the classification process. 

\subsection{Training strategy}
The proposed model utilizes a ResNet-50 architecture pre-trained on ImageNet. We modify the final fully connected layer to accommodate five primary classes of DR severity. However, this sixth class(blurred images) is only used in the loss function and does not appear in the final classification output. This approach helps the model generalize better by learning which features should be ignored, ensuring that it adapts only to the five DR severity classes. Further details on how the loss function incorporates the sixth class will be discussed in the upcoming section. The model was trained for 20 epochs with a batch size of 32 and a learning rate of 0.001. 

\subsubsection{Custom loss function}
The custom loss function combines cross-entropy loss for the five main classes with a mean squared error (MSE) loss for the sixth class. When the label corresponds to the sixth class, the model's output is compared to a uniform probability distribution across the five primary classes to encourage uncertainty and prevent reliance on irrelevant visual patterns.

The Original Image Loss function is defined as:
\begin{equation}
\mathcal{L}_{\text{OI}} = {L}_{\text{CE}}(y, \hat{y}) = -\sum_{c=1}^{C} y_c \log(\hat{y}_c)
\end{equation}
where $C$ is the number of classes, $y_c$ is the ground truth label for class $c$, and $\hat{y}_c$ is the predicted probability for class $c$.

The Blurred Image Loss for the sixth class is defined as:
\begin{equation}
\mathcal {L}_{\text{BI}} = {L}_{\text{MSE}}(\text{softmax}(\hat{y}), \mathbf{u}) = \frac{1}{C} \sum_{c=1}^{C} \left( \text{softmax}(\hat{y}_c) - u_c \right)^2
\end{equation}
where $\mathbf{u}$ is a uniform distribution vector with $u_c = 0.2$ for each class, as there are five classes.

The combined custom loss function is defined as:
\begin{equation}
\mathcal{L}_{\text{custom}} = \begin{cases}
\mathcal{L}_{\text{CE}}(y, \hat{y}) & \text{if } y \neq 5 \\
\mathcal{L}_{\text{MSE}}(\text{softmax}(\hat{y}), \mathbf{u}) & \text{if } y = 5
\end{cases}
\end{equation}

\subsection{Selecting the blur method}
\begin{figure}[ht!]
    \centering
    \includegraphics[width=0.9\textwidth]{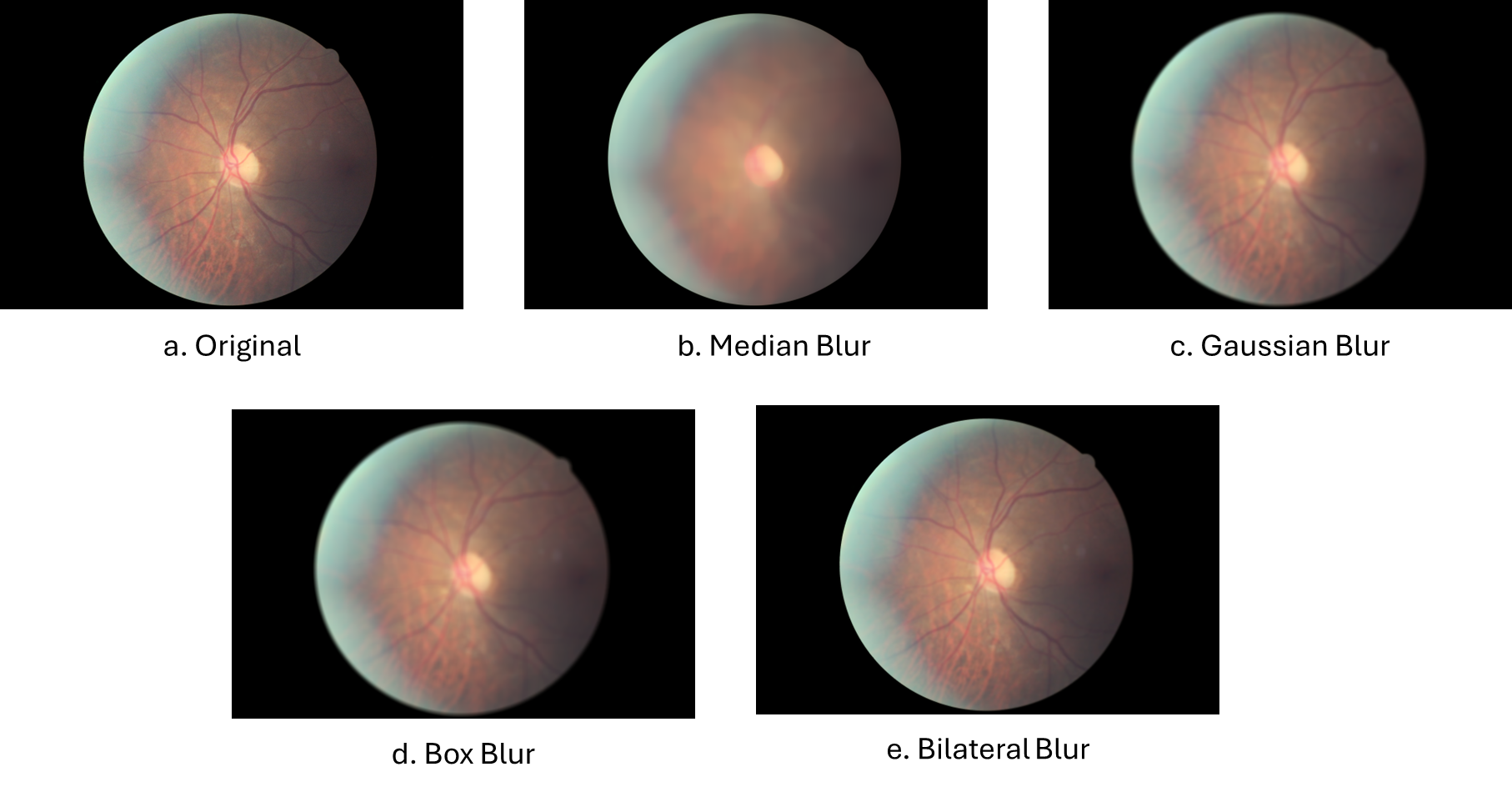}
    \caption{Examples of fundus images processed with different blur techniques. The median blur method (top center) effectively removes all blood vessels and other retinal features, leaving only the background.}
    \label{fig:blurred_methods}
\end{figure}

We experimented with several blurring techniques to determine the most effective method for removing retinal features while retaining non-informative background patterns. The considered blurring techniques are median blur, gaussian blur, box blur, and bilateral blur.

After visually inspecting the blurred images generated by each method, we found that the median blur method was the most effective at removing blood vessels and other key retinal features while retaining only the background. This ensures that the model learns to ignore non-informative background patterns during classification. Figure \ref{fig:blurred_methods} shows examples of images processed with different blur techniques, highlighting the effectiveness of the median blur in removing important features and only containing the background. Further, we did an ablation with all the blurred techniques and ensured our choice of median blur. The details of the ablation study are added under the ablation section(Section \ref{sec5}).

We conducted several experiments to evaluate the performance of our proposed method. The results are presented in the following tables, demonstrating that our method achieves superior accuracy compared to other strategies across different camera types and external datasets.

\section{Experiments and Results}\label{sec3}
In this section, we discussed different types of experiments that we followed with our approach and compared the results with the other past studies. 

\subsection{Camera type experiment}

We conducted this experiment to demonstrate that high generalization performance can be achieved without relying on camera-type information and to highlight the adversarial effects of using such information, as discussed in the RADR paper. Table~\ref{tab:camera_accuracy} presents the accuracy (\%) of different models across camera types D and E, along with their average performance under various augmentation strategies from the RADR paper. Our proposed method, AdvBlur, achieved the highest average accuracy of 82.6\%, outperforming all prior approaches. Like previous methods, we maintained consistency during training by using only cameras A, B, and C, ensuring that no camera-specific information was incorporated. These results validate the robustness of our approach in achieving superior generalization across different camera domains.

\begin{table}[ht!]
    \centering
    \caption{Performance in terms of Accuracy (\%) of our models on the test sets of the camera domains in the EyePACS dataset. SC: Single-camera training on camera A, MC: Multi-camera training on cameras A, B, and C, DA: Domain adversarial training on cameras A, B, and C. Best-performing model in bold.}
    \begin{tabular}{@{}>{\raggedright\arraybackslash}p{0.28\linewidth}>{\centering\arraybackslash}p{0.18\linewidth}>{\centering\arraybackslash}p{0.18\linewidth}>{\centering\arraybackslash}p{0.18\linewidth}@{}}
        \toprule
        Method & Camera D & Camera E & Avg \\
        \midrule
        \multicolumn{4}{c}{} \\
        SC & 67.2 $\pm$ 4.5 & 81.7 $\pm$ 0.8 & 74.5 \\
        SC ColorAug & 63.3 $\pm$ 7.3 & 80.0 $\pm$ 3.3 & 71.7 \\
        SC AugMix & 53.9 $\pm$ 3.4 & 74.9 $\pm$ 1.7 & 64.4 \\
        \midrule
        \multicolumn{4}{c}{} \\
        MC & 67.8 $\pm$ 2.7 & 82.5 $\pm$ 2.8 & 75.2 \\
        MC ColorAug & 59.1 $\pm$ 1.6 & 76.4 $\pm$ 5.7 & 67.8 \\
        MC AugMix & \underline{79.7 $\pm$ 4.6} & 83.1 $\pm$ 1.1 & 81.4 \\
        \midrule
        \multicolumn{4}{c}{} \\
        DA (RADR) & 79.1 $\pm$ 5.0 & \textbf{83.6 $\pm$ 1.4} & \underline{81.35} \\
        DA ColorAug & 60.5 $\pm$ 1.3 & 78.1 $\pm$ 6.1 & 69.3 \\
        DA AugMix & 75.4 $\pm$ 1.6 & 82.7 $\pm$ 1.3 & 79.05 \\
        \textbf{DA AdvBlur (Ours)} & \textbf{81.8 $\pm$ 0.32} & \underline{83.3 $\pm$ 0.66} & \textbf{82.6} \\
        \bottomrule
    \end{tabular}
    \label{tab:camera_accuracy}
\end{table}

\subsection{External dataset experiment}
The Table~\ref{tab:external_performance}  shows the accuracy (ACC) across different external datasets and the average performance for various training strategies, and our AdvBlur surpassed all previous work in average accuracy, which shows the effectiveness of our approach.

\begin{table}[ht!]
    \centering
    \caption{Performance of our top-performing models, MC AugMix, RADR and Ours, on the external datasets, trained with five different random seeds. SS: Single-Source training on EyePACS. MS: Multi-Source training in leave-one-out fashion on EyePACS, Messidor-1 \& 2, as well as APTOS, with prediction on the remaining dataset. Best performing model in bold, second best underlined. }
    \begin{tabular}{@{}lcccc@{}}
        \toprule
        \textbf{ACC [\%]} & Messidor-1 & Messidor-2 & APTOS & Avg \\
        \midrule
        \textbf{SS:AdvBlur (Ours)} & \underline{62.9 $\pm$ 0.41}& \textbf{74.9 $\pm$ 0.06} & \underline{68.32 $\pm$ 0.6} & \underline{68.7} \\
        SS: RADR \cite{radr2024}  & 65.3 $\pm$ 1.3 & 71.6 $\pm$ 2.2 & 60.2 $\pm$ 2.9 & 65.7\\
        SS: MC AugMix (trained by \cite{radr2024}) & 62.8 $\pm$ 2.0 & 69.8 $\pm$ 4.4 & 62.6 $\pm$ 1.4 & 65.1 \\
        SS: SPSD-ViT (trained by \cite{galappaththige2024generalizing}) & 50.5 $\pm$ 0.8 & 62.2 $\pm$ 0.4 & \textbf{75.1 $\pm$ 0.5} & 62.5 \\
        SS: DRGen \cite{galappaththige2024generalizing} & 54.6 $\pm$ 1.5 & 65.4 $\pm$ 1.1 & 61.3 $\pm$ 1.9 & 60.4 \\
        MS: SPSD-ViT \cite{galappaththige2024generalizing} & 64.8 $\pm$ 0.5 & \underline{72.4 $\pm$ 0.6} & 62.5 $\pm$ 1.2 & 69.9 \\
        MS: DANN (trained by \cite{galappaththige2024generalizing}) & 57.0 $\pm$ 1.1 & 58.6 $\pm$ 1.7 & 54.4 $\pm$ 0.8 & 56.7 \\
        MS: DRGen (trained by \cite{galappaththige2024generalizing}) & 59.1 $\pm$ 1.8 & 65.2 $\pm$ 0.6 & 51.2 $\pm$ 2.1 & 58.5 \\
        MS: DRGen \cite{atwany2022drgen} & \textbf{66.7} & 70.5 & 70.3 & \textbf{69.1} \\
        \bottomrule
    \end{tabular}
    \label{tab:external_performance}
\end{table}

The original DRGen method from \cite{atwany2022drgen} achieved the highest average accuracy of 69.1\%. Our proposed AdvBlur model secured the best performance among all single-source (SS) methods with 68.7\%, while the RADR model ranked second among SS methods with 65.7\%. However, it is important to note that this comparison favors DRGen, as their leave-one-out training and evaluation approach leveraged significantly more training data across four datasets—EyePACS, Messidor-1, Messidor-2, and APTOS. Additionally, DRGen's reported accuracies stem from different model versions per unseen dataset, whereas all our results originate from a single model, making our approach more robust and reliable for generalization.

A reproduction of DRGen under the multi-source (MS) training regime by Galappaththige et al. (2024) only achieved an average accuracy of 58.5\%, significantly lower than the original DRGen. They also evaluated DANN, a domain-adversarial network similar to our approach, under the MS regime, which only achieved 56.7\%. This is 9 percentage points lower than our single-source method, despite using more training data and multiple model instances. This suggests that our approach—avoiding camera labels and dataset-specific domain definitions—leads to better generalization than defining each dataset as a separate domain.

For a fair comparison, methods should be evaluated under the same SS training regime, where training is performed only on EyePACS and tested on all external datasets. Under these conditions, AdvBlur surpassed all SS models, outperforming RADR by 3 percentage points, SPSD-ViT by 6.2 percentage points, and an SS re-implementation of DRGen by 8.3 percentage points. These results confirm that our proposed framework competes strongly with state-of-the-art models, even when using less training data, and surpasses them under equal conditions.

Table \ref{tab:camera_accuracy} and \ref{tab:external_performance} show that our method, outperforms other approaches in experiments with different camera types and on external datasets. AdvBlur works well across different domains without relying on extra details like camera labels. It achieves better results than other studies that use such data, and importantly, it doesn't reduce the performance on the original dataset, maintaining good results in the same domain while still achieving domain generalization.

\subsection{Domain generalization results trained with smaller dataset}
To further evaluate the domain generalization capabilities of our method, we conducted single-source domain generalization experiments on various datasets. The results are presented in the following tables:

\begin{table}[ht!]
    \centering
    \caption{Single-source domain generalization results for the model trained on the Messidor-1 dataset.}
    \begin{tabular}{|l|c|c|c|c|}
        \hline
        Method      & Aptos       & Eyepacs     & Messidor-2   & Average Accuracy\\ \hline
        DRGen (trained by \cite{galappaththige2024generalizing})    & 41.7±4.3    & 43.1±7.9    & 44.8±0.9    & 43.2    \\ \hline
        AdvBlur (Ours)        & 39.1        & 58.8        & 36.6        & 44.8    \\ \hline
    \end{tabular}
    \label{tab:messidor_results}
\end{table}

\vspace{-10mm}

\begin{table}[ht!]
    \centering
    \caption{Single-source domain generalization results for the model trained on the Messidor2 dataset.}
    \begin{tabular}{|l|c|c|c|c|}
        \hline
        Method      & Aptos       & Eyepacs     & Messidor-1    & Average Accuracy\\ \hline
        DRGen (trained by \cite{galappaththige2024generalizing})    & 40.9±3.9    & 69.3±1.0    & 61.3±0.8    & 57.7    \\ \hline
        AdvBlur (Ours)        & 44.6        & 72.7        & 45.5        & 54.3    \\ \hline
    \end{tabular}
    \label{tab:messidor2_results}
\end{table}

\vspace{-10mm}

\begin{table}[ht!]
    \centering
    \caption{Single-source domain generalization results for the model trained on the Aptos dataset.}
    \begin{tabular}{|l|c|c|c|c|}
        \hline
        Method      & Eyepacs     & Messidor-1    & Messidor-2   & Average Accuracy\\ \hline
        DRGen (trained by \cite{galappaththige2024generalizing})    & 67.5±1.8    & 46.7±0.1    & 61.0±0.1    & 58.4    \\ \hline
        AdvBlur (Ours)        & 68.4        & 42.2        & 54.7        & 55.1    \\ \hline
    \end{tabular}
    \label{tab:aptos_results}
\end{table}
As we can see, our method outperformed DRGen in most of the instances, and the accuracy is higher on every dataset (Messidor-1, Messidor-2, APTOS) while trained with Messidor-1 (Table~\ref{tab:messidor_results}). However, it tends to fall behind when trained with APTOS and Messidor-2 (Table~\ref{tab:messidor2_results}, and Table~\ref{tab:aptos_results}). The results suggest that our training strategy requires a comparatively large dataset to represent each class adequately to achieve good generalization performance.

\subsection{Analyze the impact of low-quality images on our method by removing low-quality images in the training dataset (RLQI).}
This experiment evaluates the effectiveness of filtering low-quality images from the original dataset by categorizing them into three quality levels: Good, Usable, and Reject. Only images classified as Good and Usable were retained for training and testing the DR classification model. Images labeled as Reject, which exhibit severe quality issues (e.g., significant blur, uneven illumination, or low contrast), were excluded to ensure that the model was trained on diagnostically reliable data. In this approach, we employed a retinal image quality assessment (RIQA) strategy based on the method detailed by \cite{fu2019evaluation}. 

\begin{table}[ht!]
    \centering
    \caption{Accuracy (\%) across different external datasets and average performance.}
    \begin{tabular}{|>{\centering\arraybackslash}p{0.22\linewidth}|>{\centering\arraybackslash}p{0.15\linewidth}|>{\centering\arraybackslash}p{0.15\linewidth}|>{\centering\arraybackslash}p{0.15\linewidth}|>{\centering\arraybackslash}p{0.18\linewidth}|}
        \hline
        Method                  & Messidor-1 & Messidor-2 & APTOS   & Average Accuracy\\ \hline
        SS: AdvBlur (Ours)              & 62.9 ± 0.41 & 74.9 ± 0.06 & 68.0 ± 0.75 & 68.6   \\ \hline
        SS: (Ours - RLQI)       & 66.0        & 73.7        & 65.7        & 68.5   \\ \hline
    \end{tabular}
    \label{tab:external_datasets_accuracy}
\end{table}

\begin{table}[ht!]
    \centering
    \caption{Accuracy (\%) across different camera types and average performance.}
    \begin{tabular}{|>{\centering\arraybackslash}p{0.22\linewidth}|>{\centering\arraybackslash}p{0.16\linewidth}|>{\centering\arraybackslash}p{0.16\linewidth}|>{\centering\arraybackslash}p{0.18\linewidth}|}
        \hline
        Method                    & Camera D   & Camera E   & Average Accuracy\\ \hline
        DA - AdvBlur (Ours)                  & 81.8 ± 0.32 & 83.3 ± 0.66 & 82.6   \\ \hline
        DA - (Ours - RLQI)           & 82.5        & 83.5        & 83.0   \\ \hline
    \end{tabular}
    \label{tab:camera_types_accuracy}
\end{table}

As we can see from Table~\ref{tab:external_datasets_accuracy} and Table~\ref{tab:camera_types_accuracy}, RLQI does not improve the average generalization performance. This suggests that AdvBlur is not significantly affected by low-quality images. This is due to the custom loss function of AdvBlur, as the loss function itself automatically handles the effect of low-quality images in the training.

\subsection{Feature validation using Grad-CAM masking}

\begin{figure}[ht!]
    \centering
    \includegraphics[width=1\linewidth]{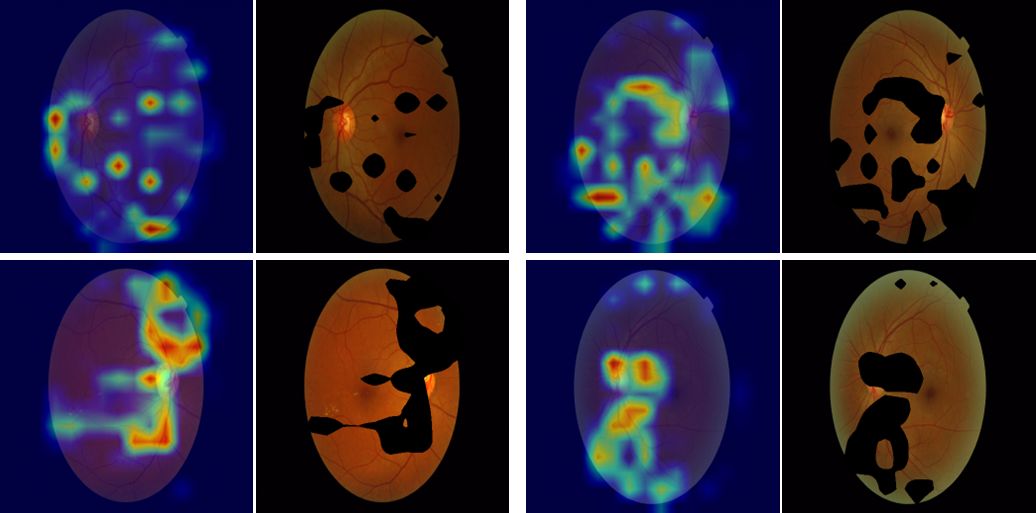}
    \caption{Heat map and the respective masked images}
    \label{fig:masked_img}
\end{figure}

To validate that our method focuses on useful features for the results, we conducted an experiment where we masked the high-activation regions identified by the Grad-CAM heatmap and re-evaluated the classification performance. The assumption was that if these regions contained essential features, occluding them would significantly degrade the model’s accuracy. Some of the masked images and the heatmaps are added in Figure \ref{fig:masked_img}. 

However, when performing classification on the masked images, we observed a notable drop in accuracy (Table \ref{tab:masking_val}), supporting that the high-activation regions identified by Grad-CAM indeed correspond to critical features for DR classification. This result provides further evidence that our model relies on meaningful and clinically relevant features when making predictions.

\begin{table}[ht!]
    \centering
    \caption{Validation accuracy (\%) for Messidor-1, Messidor-2, and APTOS with and without masking.}
    \begin{tabular}{|>{\centering\arraybackslash}p{0.25\linewidth}|>{\centering\arraybackslash}p{0.16\linewidth}|>{\centering\arraybackslash}p{0.16\linewidth}|>{\centering\arraybackslash}p{0.16\linewidth}|>{\centering\arraybackslash}p{0.18\linewidth}|}
        \hline
        Accuracy Type & Messidor-1 & Messidor-2 & APTOS & Average Accuracy \\
        \hline
        Normal Accuracy & 63.33 & 74.77 & 67.27 & 68.46 \\
        \hline
        With Masking & 55.50 & 57.62 & 63.26 & 58.79 \\
        \hline
    \end{tabular}
    \label{tab:masking_val}
\end{table}

\begin{figure}[ht!]
    \centering
    \includegraphics[width=0.4\linewidth]{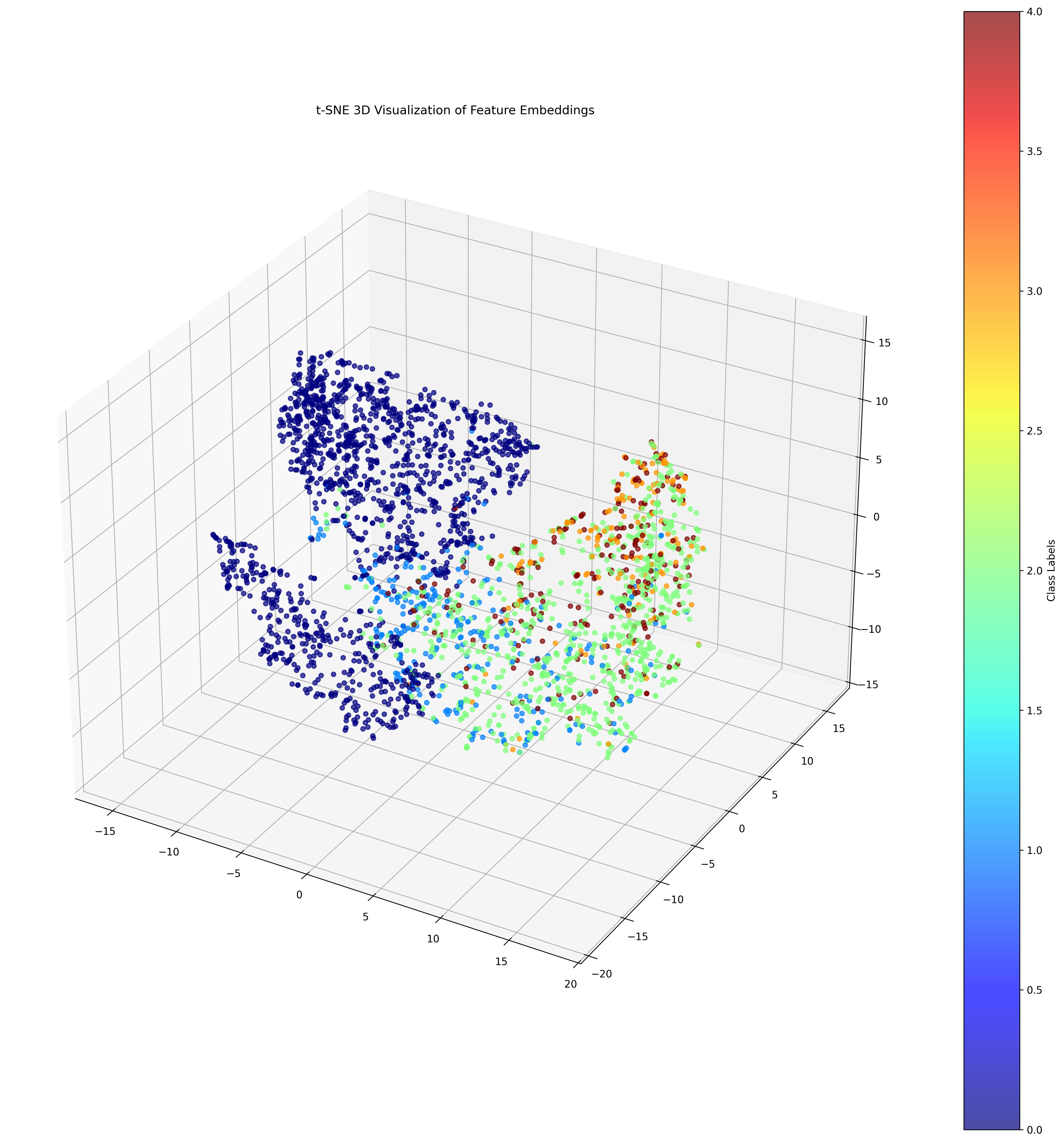}
    \caption{t-SNE plot after training- Differentiate class 0 and other classes.}
    \label{fig:tsne}
\end{figure}

\begin{figure}[ht!]
    \centering
    \includegraphics[width=0.8\linewidth]{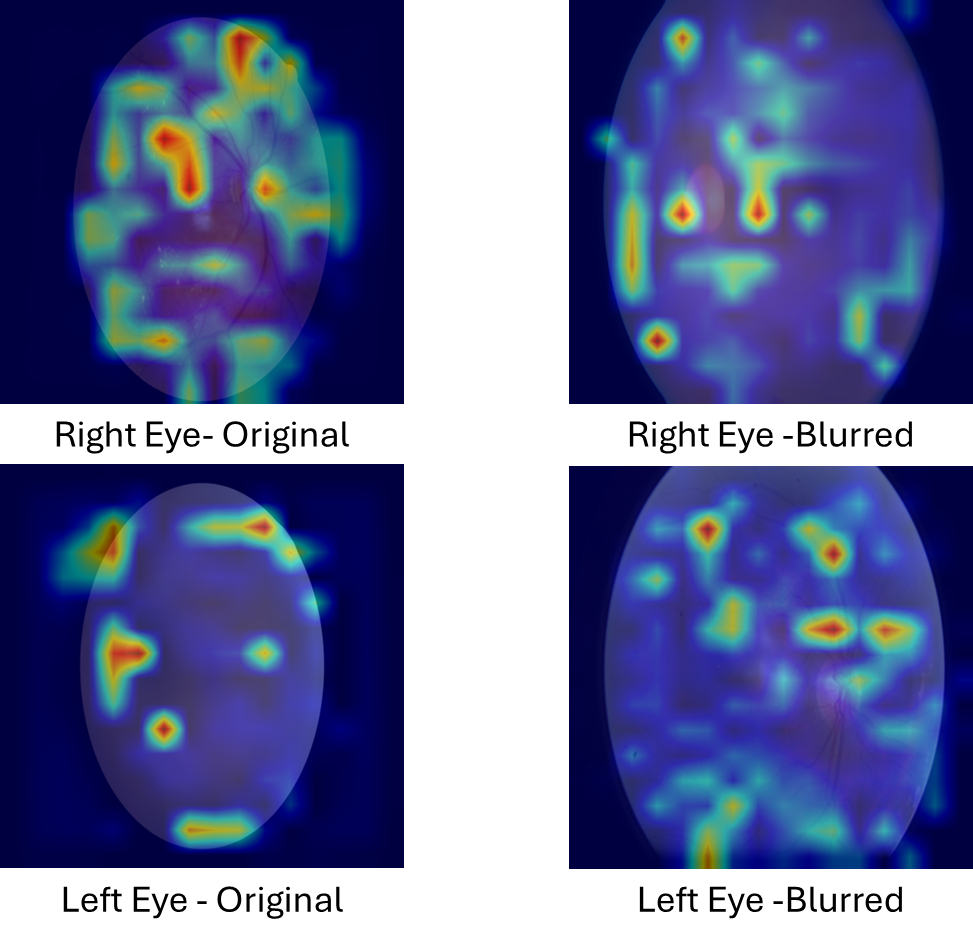}
    \caption{Gradcam Image for blurred and non-blurred images of both left and right eyes.}
    \label{fig:gradcam}
\end{figure}

We further checked our approach with the results by plotting the t-SNE (t-Distributed Stochastic Neighbor Embedding) (Figure \ref{fig:tsne}) and the GradCam (Figure \ref{fig:gradcam})
plots.

The t\_SNE plot in Figure \ref{fig:tsne} clearly shows that our approach differentiates between the label 0 (blue color) and others, which indicates that we can identify the difference between no-DR and DR.
The gradcam in Figure \ref{fig:gradcam} identifies the area of useful features, which is almost stuck in the eye region.

\section{Ablation Studies}\label{sec4}
In this section, we made several adjustments to ensure the effectiveness of our loss function and the use of blurred image techniques. 

\subsection{Ablation on loss function}
Here we trained by adding the blurred image as the 6th class, and instead of the custom loss function, we used the traditional categorical cross-entropy (CCE) loss. The accuracy of the CCE loss highlights is comparatively low compared to our custom loss function, and it shows the need for a custom loss function. The results are added in Table \ref{tab:loss_abl_1} and Table \ref{tab:loss_abl_2}.

\begin{table}[ht!]
    \centering
    \caption{Validation accuracy (\%) for Camera D and Camera E with different loss functions.}
    \begin{tabular}{|>{\centering\arraybackslash}p{0.27\linewidth}|>{\centering\arraybackslash}p{0.18\linewidth}|>{\centering\arraybackslash}p{0.18\linewidth}|>{\centering\arraybackslash}p{0.18\linewidth}|}
        \hline
        Class & DE-Class-4 & DE-Class-5 & Average Accuracy \\
        \hline
        With Custom Loss & \textbf{81.8} & \textbf{83.3} & \textbf{82.55} \\
        \hline
        With Cross-Entropy Loss & 77.40 & 83.00 & 80.20 \\
        \hline
    \end{tabular}
    \label{tab:loss_abl_1}
\end{table}

\begin{table}[ht!]
    \centering
    \caption{Validation accuracy (\%) for APTOS, Messidor-1, and Messidor-2 with different loss functions.}
    \begin{tabular}{|>{\centering\arraybackslash}p{0.27\linewidth}|>{\centering\arraybackslash}p{0.14\linewidth}|>{\centering\arraybackslash}p{0.14\linewidth}|>{\centering\arraybackslash}p{0.14\linewidth}|>{\centering\arraybackslash}p{0.18\linewidth}|}
        \hline
        Class & Aptos & Messidor-1 & Messidor-2 & Average Accuracy \\
        \hline
        With Custom Loss & \textbf{68.32} & \textbf{62.9} & \textbf{74.9} & \textbf{68.71} \\
        \hline
        With Cross-Entropy Loss & 54.22 & 74.48 & 67.75 & 65.48 \\
        \hline
    \end{tabular}
    \label{tab:loss_abl_2}
\end{table}

\subsection{Ablation on blurred images}
We did an ablation with various kinds of blurred images listed in Section\ref{sec3}. Here we followed the same AdvBlur approach as mentioned in Section\ref{sec3} and just altered the blurred techniques to ensure fair ablation. The results are shown in Table \ref{tab:blur_abl_1} \& Table \ref{tab:blur_abl_2}. The results ensure that median blur is the more prominent option as the blurred image for the approach. 
\begin{table}[ht!]
    \centering
    \caption{Validation accuracy (\%) for Camera D and Camera E, along with the average accuracy.}
    \begin{tabular}{|>{\centering\arraybackslash}p{0.27\linewidth}|>{\centering\arraybackslash}p{0.16\linewidth}|>{\centering\arraybackslash}p{0.16\linewidth}|>{\centering\arraybackslash}p{0.18\linewidth}|}
        \hline
        Blur Type & DE-Class-4 & DE-Class-5 & Average Accuracy \\
        \hline
        Median Blur & \textbf{81.8} & \textbf{83.3} & \textbf{82.55} \\
        \hline
        Gaussian Blur & 76.94 & 77.67 & 77.31 \\
        \hline
        Box Blur & 77.58 & 78.61 & 78.10 \\
        \hline
        Bilateral Blur & 78.06 & 77.34 & 77.70 \\
        \hline
    \end{tabular}
    \label{tab:blur_abl_1}
\end{table}

\begin{table}[ht!]
    \centering
    \caption{Validation accuracy (\%) for Messidor-1, Messidor-2, and APTOS, along with the average accuracy.}
    \begin{tabular}{|>{\centering\arraybackslash}p{0.27\linewidth}|>{\centering\arraybackslash}p{0.16\linewidth}|>{\centering\arraybackslash}p{0.16\linewidth}|>{\centering\arraybackslash}p{0.16\linewidth}|>{\centering\arraybackslash}p{0.18\linewidth}|}
        \hline
        Blur Type & Aptos & Messidor-1 & Messidor-2 & Average Accuracy \\
        \hline
        Median Blur & \textbf{68.32} & \textbf{62.9} & \textbf{74.9} & \textbf{68.71} \\
        \hline
        Gaussian Blur & 64.38 & 47.92 & 57.40 & 56.57 \\
        \hline
        Box Blur & 52.85 & 52.17 & 64.85 & 56.62 \\
        \hline
        Bilateral Blur & 62.83 & 51.83 & 56.48 & 57.05 \\
        \hline
    \end{tabular}
    \label{tab:blur_abl_2}
\end{table}

\section{Conclusions}\label{sec5}
DR poses a significant global health challenge, necessitating early diagnosis to prevent severe vision loss. This research detailedly analyzed domain generalization approaches in the background and identified a key gap: reliance on camera or domain labels for improved model performance. To overcome this limitation, we proposed our AdvBlur method to incorporate adversarial blurred images during training. A custom loss function was designed to encourage feature disentanglement, enabling the model to focus on critical retinal features while disregarding irrelevant domain-specific patterns.

Beyond DR, our method can be seamlessly extended to other medical imaging applications, particularly where domain generalization is crucial to mitigate ethical concerns and dataset biases.

Extensive analysis across diverse datasets and imaging conditions demonstrated that the proposed method outperforms baseline models, achieving greater robustness and diagnostic accuracy. By addressing domain generalization challenges without requiring explicit domain labels, this approach provides a scalable solution applicable to various machine vision tasks, where generalization is key.

Future work should focus on clinical deployment, integrating additional data modalities, and optimizing models for real-time applications. This research contributes to the development of robust and scalable AI-driven healthcare solutions, paving the way for more equitable and accessible medical diagnostics.

\bibliography{sn-bibliography}

\end{document}